\documentclass[english]{lni}
\begin{document}

\title[Short title]{NFDI4DS Shared Tasks for Scholarly Document Processing}

%Leuphana AIX Tilahun, Rana, Debayan, Ricardo

\author[1]{Raia Abu Ahmad}{raia.abu\string_ahmad@dfki.de}{}
\author[2]{Rana Abdulla}{rana.abdullah@leuphana.de}{}
\author[2]{Tilahun Abedissa Taffa}{tilahun.taffa@leuphana.de}{}
\author[7]{S\"{o}ren Auer}{auer@tib.eu}{}
\author[7]{Hamed Babaei Giglou}{hamed.babaei@tib.eu}{}
\author[1]{Ekaterina Borisova}{ekaterina.borisova@dfki.de}{}
\author[5]{Zongxiong Chen}
{zongxiong.chen@fokus.fraunhofer.de}{}%
\author[6]{Stefan Dietze}{stefan.dietze@gesis.org}{}
\author[7]{Jennifer D'Souza}{jennifer.dsouza@tib.eu}{}
\author[3]{Mayra Elwes}{mayra.elwes@uk-koeln.de}{%0009-0005-9454-7174
}%
\author[4]{Genet-Asefa Gesese}{genet-asefa.gesese@fiz-karlsruhe.de}{%0000-0003-3807-7145
}%
\author[4]{Shufan Jiang}{shufan.jiang@fiz-karlsruhe.de}{}%
\author[3]{Ekaterina Kutafina}{ekaterina.kutafina@uni-koeln.de}{%000-0002-3430-5123
}
\author[6]{Philipp Mayr}{philipp.mayr@gesis.org}{}
\author[1]{Georg Rehm}{georg.rehm@dfki.de}{}
\author[7]{Sameer Sadruddin}{sameer.sadruddin@tib.eu}{}
\author[5]{Sonja Schimmler}{sonja.schimmler@fokus.fraunhofer.de}{}%
\author[8]{Daniel Schneider}{daniel.schneider@uni-leipzig.de}{%0000-0003-3273-9283
}
\author[6]{Kanishka Silva}{kanishka.silva@gesis.org}{}
\author[6]{Sharmila Upadhyaya}{sharmila.upadhyaya@gesis.org}{}
\author[2]{Ricardo	Usbeck}{ricardo.usbeck@leuphana.de}{}

\affil[1]{Deutsches Forschungszentrum für Künstliche Intelligenz GmbH (DFKI), Berlin, Germany}
\affil[2]{Leuphana University of Lüneburg, Germany}
\affil[3]{University Hospital Cologne, University of Cologne, Institute for Biomedical Informatics, Germany}
\affil[4]{FIZ-Karlsruhe – Leibniz-Institute for Information Infrastructure, Germany}
\affil[5]{Fraunhofer Institute for Open Communication Systems FOKUS, Berlin, Germany}
\affil[6]{GESIS -- Leibniz Institute for the Social Sciences, Cologne, Germany}
\affil[7]{TIB Leibniz Information Centre for Science and Technology, Hannover, Germany}
\affil[8]{Innovation Center Computer Assisted Surgery (ICCAS), Leipzig University, Germany}
 
\maketitle

\begin{abstract}
Shared tasks are powerful tools for advancing research through community-based standardised evaluation. As such, they play a key role in promoting findable, accessible, interoperable, and reusable (FAIR), as well as transparent and reproducible research practices. This paper presents an updated overview of twelve shared tasks developed and hosted under the German National Research Data Infrastructure for Data Science and Artificial Intelligence (NFDI4DS) consortium, covering a diverse set of challenges in scholarly document processing. Hosted at leading venues, the tasks foster methodological innovations and contribute open-access datasets, models, and tools for the broader research community, which are integrated into the consortium's research data infrastructure. 
\end{abstract}
%%% Beginn des Artikeltexts

\section{Introduction}

Shared tasks are community-driven challenges that compare computational methods on specific problems using shared datasets and metrics \cite{parra-escartin-etal-2017-ethical}. They advance research by fostering innovation, encouraging collaboration, and establishing state-of-the-art methods across tasks and fields \cite{filannino2018advancing}. Shared tasks offer multiple benefits to the community by promoting transparency and reproducibility, allowing for a fair comparison of methodologies, and resulting in publicly available resources (e.\,g., datasets, open-source software) \cite{escartin2021towards}.

NFDI4DS (National Research Data Infrastructure for Data Science and Artificial Intelligence) is a German consortium that focuses on supporting the research data lifecycle in Data Science and Artificial Intelligence (AI) \cite{schimmler2023nfdi4ds}.\footnote{\url{https://www.nfdi4datascience.de}} This includes collecting, processing, analysing, publishing, and reusing artifacts such as datasets, models, code, and publications. NFDI4DS pays special attention to the FAIR principles as well as to transparency and reproducibility through open-source tools and rich documentation.

Shared tasks play an important role in NFDI4DS, serving as mechanisms for community engagement and driving the development and refinement of tools and methods for its FAIR research data infrastructure. They also support NFDI’s broader mission of building a sustainable, interoperable, and accessible national research data infrastructure by promoting interdisciplinary collaboration, standardisation efforts, and long-term reuse of research outputs. %The resulting research artifacts and digital objects are integrated into the broader NFDI4DS infrastructure, ensuring their sustainability and accessibility.

To reach a broader audience and encourage widespread participation, NFDI4DS shared tasks are usually hosted at well-known venues. Currently, these include 1.~the Natural Scientific Language Processing and Research Knowledge Graphs (NSLP) workshop series, initiated by NFDI4DS;\footnote{\url{https://nfdi4ds.github.io/nslp2025/} and \citet{Rehm2024}} 2.~the Scholarly Document Processing (SDP) workshop series \cite{SDP2020};\footnote{\url{https://sdproc.org}} 3.~the Conference and Labs of the Evaluation Forum (CLEF);\footnote{\url{https://clef2025.clef-initiative.eu/}} 4.~the Workshop on Semantic Evaluation (SemEval);\footnote{\url{https://semeval.github.io}} and 5.~the Int.~Semantic Web Conference (ISWC) challenges.\footnote{\url{https://iswc2025.semanticweb.org/\#/program/challenges}}  

This paper provides an overview of shared tasks organised by NFDI4DS, extending \cite{abu2023nfdi4ds}. The twelve tasks demonstrate how these competitions can drive methodological progress and contribute to the development of a FAIR research data infrastructure.

\section{Completed Shared Tasks}

\subsection{SOMD 2025: Software Mention Detection}
%GESIS

\textbf{Organiser: GESIS -- Leibniz Institute for the Social Sciences} \\
Although software is integral to scientific discovery and analysis, it is rarely cited with the same consistency or formality as literature. Automatic identification and disambiguation of software mentions and related attributes (URLs, Versions, License, etc.) enhance accessibility and reproducibility but remain challenging \cite{somsci}. SOMD 2025 \cite{somd-2025} is a shared task focused on extracting software mentions from scholarly articles. Following SOMD 2024 \cite{franketal}, the task focused on joint learning of software mentions, attribute extraction, and classification of the their relations. The task was hosted at SDP 2025,\footnote{\url{https://sdproc.org/2025/somd25.html}} and included two phases: Phase I involved model development using a gold standard training set \cite{somsci}, evaluated using the average score of the entities extraction and relation classification tasks, while phase II tested generalisability on an out-of-distribution set. The competition (March/April 2025) attracted six teams, the best one achieved F1 scores of 0.89 (Phase I) and 0.63 (Phase II), surpassing baselines of 0.804 and 0.491, respectively.

\subsection{SciVQA: Scientific Visual Question Answering}

\textbf{Organiser: Deutsches Forschungszentrum für Künstliche Intelligenz (DFKI)} \\
The first SciVQA shared task \cite{borisova-scivqa-2025}, part of SDP 2025,\footnote{\url{https://sdproc.org/2025/scivqa.html}} is about developing question answering (QA) systems based on images of scientific figures, their captions, and metadata. The challenge aims to evaluate the recognition and reasoning capabilities of multimodal large language models (LLMs) on closed-ended visual vs.~non-visual questions. The former refers to questions addressing visual attributes such as colour, shape, direction, height, size or position of objects, while the latter does not incorporate any of these aspects. The novel and large-scale SciVQA dataset\footnote{\url{https://huggingface.co/datasets/katebor/SciVQA}} comprises 3000 images of figures from arXiv pre-prints and ACL Anthology publications. Each figure is semi-automatically annotated according to a custom QA schema featuring seven types, resulting in a total of 21000 English QA pairs. The benchmark contains several types of figures (e.\,g., line graph, bar chart, venn diagram, confusion matrix), also distinguishing between compound and non-compound figures. The evaluation is based on precision, recall and F1-scores of ROUGE-1, ROUGE-L \cite{lin-2004-rouge}, and BertScore \cite{zhang2020BERTScore}.\footnote{\url{https://www.codabench.org/competitions/5904/}} The competition attracted 20 registrations, with seven leaderboard submissions and five papers reporting their solutions. The highest-ranking team achieved scores of 0.80 for ROUGE-1 and ROUGE-L and 0.98 for BertScore. 

\subsection{ClimateCheck: Scientific Fact-checking of Social Media Posts on Climate Change}

\textbf{Organiser: Deutsches Forschungszentrum für Künstliche Intelligenz (DFKI)} \\
ClimateCheck \cite{climatecheck-shared-task} was hosted at SDP 2025.\footnote{\url{https://sdproc.org/2025/climatecheck.html}} The task addressed the growing challenge of misinformation in online discourse about critical issues, taking climate change as an example, by linking claims made on social media to relevant scholarly articles. Two datasets were constructed for the task \cite{climatecheck-dataset}, the first consists of English climate-related claims gathered from existing resources,  while the second is a corpus of 394,269 publications in climate and environmental sciences. The claims were filtered for scientific check-worthiness and rephrased to be atomic tweet-styled statements. A TREC-like \cite{voorhees2005trec} linking approach was then used to create claim-abstract pairs for annotation, resulting in a dataset of 435 unique claims, each connected to abstracts annotated as \emph{supports}, \emph{refutes}, or \emph{not enough information}.\footnote{\url{https://huggingface.co/datasets/rabuahmad/climatecheck}} Subtask I of ClimateCheck focused on retrieving the top 10 relevant abstracts per claim, evaluated using Recall@K as well as Binary Preference (Bpref), while subtask II asked participants to classify the relation of each claim-abstract pair they retrieved, evaluated using F1-score with additional scaling based on successful retrieval.
The task was hosted on CodaBench \cite{codabench}, attracting 27 participants with ten leaderboard submissions. The top team achieved a Recall@10 score of 0.66, Bpref of 0.49 (subtask I), and and F1 of 0.73 (subtask II). 

\subsection{FoRC2025: Field of Research Classification}
%DFKI (Raia)

\textbf{Organiser: Deutsches Forschungszentrum für Künstliche Intelligenz (DFKI)} \\
The second FoRC shared task \cite{francis2025overview} was hosted at NSLP 2025,\footnote{\url{https://nfdi4ds.github.io/nslp2025/docs/forc_shared_task.html}} running in February/April 2025 on CodaBench. Building on FoRC 2024 \cite{abu2024forc}, the 2025 iteration reused the FoRC4CL dataset with 1,500 ACL Anthology papers annotated using a taxonomy of 170 topics \cite{ahmad2024forc4cl}. For 2025, a weakly-labeled dataset of 41,000 automatically annotated articles was added. Given an ACL Anthology article, the task was to predict all labels depicting its main contributions. While only one team submitted results, they significantly outperformed the baselines and previous systems using a two-stage approach: k-nearest neighbors for label retrieval, followed by zero-shot prompting for refinement. Their model yielded a micro-F1 of 0.68, a macro-F1 of 0.66, and a weighted-F1 of 0.69, showing the potential of hybrid retrieval-generation pipelines. The FoRC4CL 2025 dataset offers a valuable benchmark for future research on hierarchical multi-label classification, weak supervision methods, and the integration of LLMs in scientific document understanding.  

\subsection{LLMs4Subjects@SemEval-2025: LLM-based Automated Subject Tagging for a National Technical Library's Open-Access Catalog}

%TIB (Jennifer/Sameer)
%author full names to be credited in the paper
% Jennifer D'Souza, TIB Leibniz Information Centre for Science and Technology, Hannover, Germany
% Sameer Sadruddin, TIB Leibniz Information Centre for Science and Technology, Hannover, Germany

\textbf{Organiser: TIB Leibniz Information Centre for Science and Technology} \\
The first edition of LLMs4Subjects, organised as SemEval-2025 Task 5,\footnote{\url{https://semeval.github.io/SemEval2025/tasks.html}} focused on automated subject tagging of scientific and technical records using LLMs \cite{d2025semeval}. The task targeted records in English and German from the TIBKAT catalog, Germany’s national technical library (5.7 million records). The dataset comprised 123,589 open-access records annotated with subject terms from the Gemeinsame Normdatei (GND) \cite{llms4subjects-dataset}. Two variants were offered: (1) the \textit{all-subjects} collection ($\sim$100k records, full GND coverage) and (2) the \textit{tib-core} subset ($\sim$48k records, 14 key technical domains), each accompanied by a refined GND taxonomy. In total, 14 teams participated with approaches ranging from extreme multi-label text classification with LLM augmentation (e.\,g., \cite{annif}), to few-shot LLM ensembles \cite{dnb-ai}, to retrieval-augmented generation (e.\,g., \cite{homa}). Quantitative and qualitative evaluations were conducted, with subject specialists reviewing model outputs. Key insights included the benefits of synthetic data generation, LLM-assisted translation, multilingual model design, and contrastive learning.\footnote{\url{https://sites.google.com/view/llms4subjects/team-results-leaderboard}} The task established a strong foundation for scalable, bilingual subject tagging in digital libraries. The next edition\footnote{\url{https://sites.google.com/view/llms4subjects-germeval/}} will focus on \textit{energy- and compute-efficient} LLM systems as part of the GermEval workshop series.

\subsection{LLMs4OL-2024: Large Language Models for Ontology Learning}

%TIB (Hamed/Jennifer/Soeren)
%author full names to be credited in the paper
% Hamed Babaei Giglou, TIB Leibniz Information Centre for Science and Technology, Hannover, Germany
% Jennifer D'Souza, TIB Leibniz Information Centre for Science and Technology, Hannover, Germany
% S\"{o}ren Auer, TIB Leibniz Information Centre for Science and Technology, Hannover, Germany & L3S Research Center, Leibniz University of Hannover, Germany

\textbf{Organiser: TIB Leibniz Information Centre for Science and Technology} \\
The first LLMs4OL challenge \cite{giglou2024llms4ol} was organised at ISWC 2024 and co-supported by the SCINEXT project. The goal was to evaluate and advance the use of LLMs for ontology learning (OL) tasks, including term typing, taxonomy induction, and non-taxonomic relation extraction \cite{babaei2023llms4ol}. The challenge comprised three tasks and 21 subtasks, using datasets derived from ontologies such as WordNet, GeoNames, and UMLS. Datasets were annotated for both few-shot and zero-shot setups, allowing for robust evaluation of model generalisation \cite{llms4ol_dataset}. The challenge attracted 14 participating teams on Codalab,\footnote{\url{https://codalab.lisn.upsaclay.fr/competitions/19547}} eight of which reached the final leaderboard. Winning approaches combined traditional machine learning with LLM prompt-tuning, rules-based augmentation, or retrieval-enhanced methods. The diversity of methods and breadth of domains addressed established a strong benchmark for future research, highlighting the importance of hybrid strategies and domain adaptation for effective ontology construction. A second edition, LLMs4OL 2025,\footnote{\url{https://sites.google.com/view/llms4ol2025}} is planned to be co-located with ISWC 2025. A major addition is a new task, \textit{text2onto}, which evaluates systems for their ability to generate structured ontologies from unstructured text.

\subsection{Scholarly QALD: Scholarly Question Answering over Linked Data}

\textbf{Organiser: Leuphana University of Lüneburg, Germany} \\
The QALD challenge addresses the development of natural language interfaces over scholarly knowledge graphs (KGs), enabling users to access and query them without requiring expertise in semantic technologies. It was held at ISWC 2023 and ISWC 2024, aiming to foster the development and evaluation of QA systems tailored to scholarly data. The 2023 edition comprised two main tasks: 1.~DBLP-QUAD (KGQA over DBLP), in which participants worked with a dataset of 10,000 question-SPARQL pairs, designed for queries over DBLP; and 2.~SciQA (KGQA over ORKG), which focused on answering comparative scholarly questions over the ORKG. The task attracted seven teams \cite{QALD_SemREC_2023} and was evaluated using the F1 score.\footnote{\url{https://kgqa.github.io/scholarly-QALD-challenge/2023/}} In 2024, a new Hybrid Question Answering (Hybrid-QA) task was introduced, requiring participants to fetch answers by querying multiple scholarly KGs and textual resources, reflecting more realistic, multi-source user needs in the scholarly domain \cite{taffa2024hybrid}. Three teams submitted results, and the evaluation was expanded to a broader set of metrics, including Exact Match, F-Score, and METEOR.\footnote{\url{https://kgqa.github.io/scholarly-QALD-challenge/2024/}} 

\section{Ongoing Shared Tasks}

\subsection{CheckThat!Lab Task 4: Scientific Web Discourse}

\textbf{Organiser: GESIS -- Leibniz Institute for the Social Sciences} \\
Scientific web discourse has increased substantially throughout the past years \cite{dunwoody2021science,bruggemann2020post}. It is usually informal and exhibits incomplete citation habits, where the actual study is never explicitly cited. This poses challenges both from a computational perspective, when mining social media or computing Altmetrics, and a societal perspective, leading to poorly informed online debates \cite{rocha2021impact}. To address this, two tasks are introduced as part of the CLEF 2025 CheckThat!~Lab.\footnote{\url{https://checkthat.gitlab.io/clef2025/}} Subtask 4.1, SciWeb Discourse Detection, aims to classify the different forms of science-related online discourse  \cite{hafid2022scitweets}. Given a tweet, the task is to detect if it contains a scientific claim, a scientific reference, or is referring to science contexts or entities. The SciTweets corpus \cite{hafid2022scitweets} will be used with an additional test set of 500-1000 tweets. Systems will be evaluated using macro-F1. Subtask 4.2, SciWeb Claim-Source Retrieval, aims to retrieve the scientific paper that serves as the source for a claim with an implicit scientific reference. A ground truth dataset containing tweet-study pairs will be provided, using the CORD-19 corpus \cite{wang-etal-2020-cord}. Tweets are collected using the Altmetric corpus\footnote{\url{https://www.altmetric.com}} and a pretrained classifier \cite{hafid2022scitweets}, resulting in a collection of 15,967 tweet-study pairs. Systems will be evaluated using Mean Reciprocal Rank (MRR@5). 

\begin{comment}
We collect scientific papers from the CORD-19 corpus \cite{wang-etal-2020-cord}, where each data point contains the title, abstract, and metadata of a paper related to COVID-19 research.
Using the Altmetric corpus\footnote{\url{https://www.altmetric.com}} and a pretrained classifier \cite{hafid2022scitweets}, we collect tweets that contain both scientific claims and explicit study references, where explicit references such as URIs will be removed from the test set. Our final collection comprises 15,967 tweet-study pairs, where each tweet points to an academic paper from the CORD-19 corpus.
We will use the Mean Reciprocal Rank score (\textit{MRR@5}) to evaluate submissions to this task.
\end{comment}

\section{Shared Tasks Under Development}

\subsection{Dataset and Machine Learning Model Mention Detection}

\textbf{Organiser: GESIS -- Leibniz Institute for the Social Sciences} \\
Machine Learning (ML) models and datasets are the foundation of scientific discovery and innovation in AI. Their occurrence in scholarly articles is crucial for understanding and reproducing of research. With the intent of creating novel tasks and corpora of scholarly work to contribute to the fine-grained representation of research work, the GSAP-NER corpora \cite{otto2023gsapnernoveltaskcorpus} were designed. This set of gold-standard datasets provides a nuanced understanding of how ML models and datasets are mentioned and utilised in scholarly articles. The shared task will use GSAP-NER and encourage researchers to work on novel solutions for detecting ML models and datasets, optimising the automated extraction of these entities from scholarly articles. 

%\section{MESD: Metadata Extraction from Scholarly Documents}
%Fraunhofer FIT (Zeyd)

\subsection{ReadMe2KG: Github ReadMe to Knowledge Graph}

\textbf{Organiser: FIZ Karlsruhe - Leibniz Institute for Information Infrastructure} \\
GitHub\footnote{\url{https://github.com}} is a popular platform for hosting and collaborating on software projects. Researchers use GitHub repositories to share datasets, models, and the source code of their experiments. These repositories can provide implementation details and facilitate the exploration and reproduction of research results, each  typically including a README.md file, which serves as an introductory document. READMEs are usually written in Markdown format and provide key information such as the project’s purpose, setup instructions, usage examples, and often links to the original research paper. Aiming to enhance the NDFI4DS-KG\footnote{\url{1https://nfdi.fiz-karlsruhe.de/4ds/sparql}, \url{https://nfdi.fiz-karlsruhe.de/4ds/shmarql}} with information from GitHub README files, two subtasks are proposed. In Subtask I, Fine-grained Named Entity Recognition, participants will develop classifiers that take README files as input and output the mentions of the $\sim$15 classes in the NFDI4DS Ontology \cite{gesese2024nfdi4dso}. A dataset with $\sim$160 README.md files will be made available to train classifiers, which will be evaluated using micro and macro scores of recall, precision, and F1. Subtask II, Entity Linking, will task participants with developing a method to link entities mentioned in README files to entities in the NFDI4DS-KG. A dataset with  $\sim$200 README.md files will be made available. The systems will be evaluated using micro and macro scores of recall, precision, and F1.

\subsection{CURED: Clinical Uncertainty and Reliable Evidence-based Decision-making}
% ULEI
% \author[x]{Daniel Schneider}{daniel.schneider@uni-leipzig.de}{0000-0003-3273-9283}
% \affil[x]{Innovation Center Computer Assisted Surgery (ICCAS), Leipzig University, Germany}

\textbf{Organiser: Innovation Center Computer Assisted Surgery (ICCAS)} \\
CURED is a planned shared task on trustworthy, evidence-based clinical prognosis of patient trajectories. In safety‐critical settings like clinical decision-making, AI applications must operate under expert supervision. Clinical decision support (CDS) systems fulfill this role by processing large volumes of data to generate actionable insights, while ensuring that clinicians retain final control. However, AI‐driven CDS models frequently exhibit overconfidence when faced with outliers or distribution shifts \cite{guo_2017}. To foster reliable human–AI collaboration and support evidence‐based decisions, CDS systems must transparently communicate their uncertainty and limits of their knowledge \cite{lindenmeyer_2025}. However, no adequate standardised benchmark currently exists for evaluating knowledge uncertainty estimates in clinical settings. To address this gap, we propose CURED, a shared task to evaluate collaborative clinical decision making in multi-agent settings by leveraging uncertainty quantification to efficiently integrate each agent’s expertise. CURED will challenge participants to build prognostic models on longitudinal electronic health records that output both a prediction and calibrated epistemic confidence estimate. Submissions will be ranked according to both discriminative performance and the quality of their uncertainty estimates. CURED is currently in the dataset creation phase, with a plan to run in 2026.

\subsection{Algorithm2Domain}

\textbf{Organiser: University Hospital Cologne, University of Cologne} \\
Reusability of data science algorithms is crucial for efficient and scalable research. To maximise their generalisability to unseen data and adaptability across diverse distributions, promoting best-practices in algorithm development is essential. Simultaneously, new projects need effective ways to discover existing algorithms that generalise well to their specific data. To address this, we initiated the development of a meta-benchmarking platform to evaluate algorithm generalisability and test domain adaptation methods. The concept is built on four key pillars: 1.~Decomposition of algorithm development into ``degrees of freedom'' for modular benchmarking of computational components; 2.~Collaborative development of best-practice benchmarking guidelines; 3.~Establishment of FAIR standards for benchmarking, drawing on existing frameworks such as Croissant\footnote{\url{https://github.com/mlcommons/croissant}}; 4.~Co-creation of a shared knowledge base on algorithm generalisation and adaptation. Algorithm2Domain,\footnote{\url{https://github.com/BI-K/Algorithm2Domain/}} builds on insights from the ADATIME benchmarking suite \cite{ragab2023adatime},
mapping its structure to the ``degrees of freedom'' framework, testing the process of adding new modules, and exploring no-code GUI solutions for rapid testing. We aim to establish a development roadmap for these pillars and build the infrastructure to support community-driven contributions in shared tasks, such as the preparation of algorithms, benchmarking, and expansion of the collaborative knowledge base.

\section{Conclusion}

This paper presented a total of twelve shared tasks, all of which are part of the NFDI4DS consortium, which aims to contribute a FAIR, transparent, and reproducible research data infrastructures. The presented shared tasks were split into seven completed, one ongoing, and four future tasks, demonstrating the consortium's efforts in tackling challenging problems in the scholarly information processing domain using a community-based approach. These efforts directly contribute to the broader NFDI mission of enabling sustainable and interoperable research data practices across different disciplines.

\printbibliography

@incollection{dunwoody2021science,
  title={Science journalism: Prospects in the digital age},
  author={Dunwoody, Sharon},
  booktitle={Routledge handbook of public communication of science and technology},
  pages={14--32},
  year={2021},
  publisher={Routledge}
}

@article{bruggemann2020post,
  title={Post-normal science communication: exploring the blurring boundaries of science and journalism},
  author={Br{\"u}ggemann, Michael and L{\"o}rcher, Ines and Walter, Stefanie},
  journal={Journal of Science Communication},
  volume={19},
  number={3},
  pages={A02},
  year={2020},
  publisher={SISSA Medialab srl}
}

@inproceedings{guo_2017,
    author = {Guo, Chuan and Pleiss, Geoff and Sun, Yu and Weinberger, Kilian Q.},
    title = {On calibration of modern neural networks},
    year = {2017},
    publisher = {JMLR.org},
    booktitle = {Proceedings of the 34th International Conference on Machine Learning - Volume 70},
    pages = {1321–1330},
    numpages = {10},
    location = {Sydney, NSW, Australia},
    series = {ICML'17}
}

@article{lindenmeyer_2025,
    AUTHOR = {Lindenmeyer, Adrian and Blattmann, Malte and Franke, Stefan and Neumuth, Thomas and Schneider, Daniel},
    TITLE = {Towards Trustworthy AI in Healthcare: Epistemic Uncertainty Estimation for Clinical Decision Support},
    JOURNAL = {Journal of Personalized Medicine},
    VOLUME = {15},
    NUMBER = {2},
    ARTICLE-NUMBER = {58},
    PubMedID = {39997335},
    ISSN = {2075-4426},
    DOI = {10.3390/jpm15020058},
    YEAR = {2025},

}

@article{rocha2021impact,
  title={The impact of fake news on social media and its influence on health during the COVID-19 pandemic: A systematic review},
  author={Rocha, Yasmim Mendes and de Moura, Gabriel Ac{\'a}cio and Desid{\'e}rio, Gabriel Alves and de Oliveira, Carlos Henrique and Louren{\c{c}}o, Francisco Dantas and de Figueiredo Nicolete, Larissa Deadame},
  journal={Journal of Public Health},
  pages={1--10},
  year={2021},
  publisher={Springer}
}

@inproceedings{hafid2022scitweets,
  title={SciTweets-A Dataset and Annotation Framework for Detecting Scientific Online Discourse},
  author={Hafid, Salim and Schellhammer, Sebastian and Bringay, Sandra and Todorov, Konstantin and Dietze, Stefan},
  booktitle={Proceedings of the 31st ACM International Conference on Information \& Knowledge Management},
  pages={3988--3992},
  year={2022}
}

@inproceedings{wang-etal-2020-cord,
    title = "{CORD-19}: The {COVID-19} Open Research Dataset",
    author = "Wang, Lucy Lu  and Lo, Kyle  and Chandrasekhar, Yoganand  and Reas, Russell  and Yang, Jiangjiang  and Burdick, Doug  and Eide, Darrin  and Funk, Kathryn  and Katsis, Yannis  and Kinney, Rodney Michael  and Li, Yunyao  and Liu, Ziyang  and Merrill, William  and Mooney, Paul  and Murdick, Dewey A.  and Rishi, Devvret  and Sheehan, Jerry  and Shen, Zhihong  and Stilson, Brandon  and Wade, Alex D.  and Wang, Kuansan  and Wang, Nancy Xin Ru  and Wilhelm, Christopher  and Xie, Boya  and Raymond, Douglas M.  and Weld, Daniel S.  and Etzioni, Oren  and Kohlmeier, Sebastian",
    booktitle = "Proceedings of the 1st Workshop on {NLP} for {COVID-19} at {ACL} 2020",
    month = jul,
    year = "2020",
    address = "Online",
    publisher = "Association for Computational Linguistics",
}

@inproceedings{franketal,
author = {Kr\"{u}ger, Frank and Karmakar, Saurav and Dietze, Stefan},
title = {SOMD@NSLP2024: Overview and Insights from the Software Mention Detection Shared Task},
year = {2024},
isbn = {978-3-031-65793-1},
publisher = {Springer-Verlag},
address = {Berlin, Heidelberg},
doi = {10.1007/978-3-031-65794-8_17},
booktitle = {Natural Scientific Language Processing and Research Knowledge Graphs: First International Workshop, NSLP 2024, Hersonissos, Crete, Greece, May 27, 2024, Proceedings},
pages = {247–256},
numpages = {10},
location = {Hersonissos, Crete, Greece}
}

@inproceedings{somsci,
author = {Schindler, David and Bensmann, Felix and Dietze, Stefan and Kr\"{u}ger, Frank},
title = {SoMeSci- A 5 Star Open Data Gold Standard Knowledge Graph of Software Mentions in Scientific Articles},
year = {2021},
isbn = {9781450384469},
publisher = {Association for Computing Machinery},
address = {New York, NY, USA},
doi = {10.1145/3459637.3482017},
booktitle = {Proceedings of the 30th ACM International Conference on Information \& Knowledge Management},
pages = {4574–4583},
numpages = {10},
location = {Virtual Event, Queensland, Australia},
series = {CIKM '21}
}

@Book{Rehm2024,
  editor = {Georg Rehm and Stefan Dietze and Sonja Schimmler and Frank
                  Krüger},
  title = {Natural Scientific Language Processing and Research
                  Knowledge Graphs (NSLP 2024): First International Workshop},
  publisher = {Springer},
  year = {2024},
  number = {14770},
  series = {Lecture Notes in Artificial Intelligence (LNAI)},
  address = {Cham, Switzerland},
  http = {https://link.springer.com/book/10.1007/978-3-031-65794-8},
  doi = {https://doi.org/10.1007/978-3-031-65794-8},
  note = {Hersonissos, Crete, Greece, 27 May 2024}
}

@inproceedings{SDP2020,
    title = "Overview of the First Workshop on Scholarly Document Processing ({SDP})",
    author = "Chandrasekaran, Muthu Kumar  and
      Feigenblat, Guy  and
      Freitag, Dayne  and
      Ghosal, Tirthankar  and
      Hovy, Eduard  and
      Mayr, Philipp  and
      Shmueli-Scheuer, Michal  and
      de Waard, Anita",
     booktitle = "Proceedings of the First Workshop on Scholarly Document Processing",
    month = nov,
    year = "2020",
    address = "Online",
    publisher = "Association for Computational Linguistics",
    doi = "10.18653/v1/2020.sdp-1.1",
    pages = "1--6",
}

@proceedings{QALD_SemREC_2023,
  editor       = {Debayan Banerjee and Ricardo Usbeck and Nandana Mihindukulasooriya and Mohamad Yaser Jaradeh and Sören Auer and Gunjan Singh and Raghava Mutharaju and Pavan Kapanipathi},
  title        = {Joint Proceedings of Scholarly QALD 2023 and SemREC 2023: 1st Scholarly QALD Challenge 2023 and 4th SeMantic Answer Type, Relation and Entity Prediction Tasks Challenge 2023},
  booktitle    = {22nd International Semantic Web Conference (ISWC 2023)},
  address      = {Athens, Greece},
  month        = nov,
  year         = {2023},
  publisher    = {CEUR Workshop Proceedings},
  volume       = {3592},
}

@article{taffa2024hybrid,
  title={Hybrid-SQuAD: Hybrid Scholarly Question Answering Dataset},
  author={Taffa, Tilahun Abedissa and Banerjee, Debayan and Assabie, Yaregal and Usbeck, Ricardo},
  journal={arXiv preprint arXiv:2412.02788},
  year={2024}
}

@inproceedings{parra-escartin-etal-2017-ethical,
    title = "Ethical Considerations in {NLP} Shared Tasks",
    author = "Parra Escart{\'i}n, Carla  and
      Reijers, Wessel  and
      Lynn, Teresa  and
      Moorkens, Joss  and
      Way, Andy  and
      Liu, Chao-Hong",
    editor = "Hovy, Dirk  and
      Spruit, Shannon  and
      Mitchell, Margaret  and
      Bender, Emily M.  and
      Strube, Michael  and
      Wallach, Hanna",
    booktitle = "Proceedings of the First {ACL} Workshop on Ethics in Natural Language Processing",
    month = apr,
    year = "2017",
    address = "Valencia, Spain",
    publisher = "Association for Computational Linguistics",
    doi = "10.18653/v1/W17-1608",
    pages = "66--73",
}

@article{filannino2018advancing,
  title={Advancing the State of the Art in Clinical Natural Language Processing through Shared Tasks},
  author={Filannino, Michele and Uzuner, {\"O}zlem},
  journal={Yearbook of medical informatics},
  volume={27},
  number={1},
  pages={184--192},
  year={2018}
}

@article{schimmler2023nfdi4ds,
  title={NFDI4DS infrastructure and services},
  author={Schimmler, Sonja and Wentzel, Bianca and Bleier, Arnim and Dietze, Stefan and Karmakar, Saurav and Mutschke, Peter and Kraft, Angelie and Taffa, Tilahun A and Usbeck, Ricardo and Boukhers, Zeyd and others},
  year={2023},
  journal={Gesellschaft f{\"u}r Informatik eV}
}

@article{escartin2021towards,
  title={Towards transparency in NLP shared tasks},
  author={Escart{\'\i}n, Carla Parra and Lynn, Teresa and Moorkens, Joss and Dunne, Jane},
  journal={arXiv preprint arXiv:2105.05020},
  year={2021}
}

@article{abu2023nfdi4ds,
  title={NFDI4DS Shared Tasks},
  author={Abu Ahmad, Raia and Borisova, Ekaterina and Rehm, Georg and Dietze, Stefan and Kamakar, Saurav and Otto, Wolfgang and D’Souza, Jennifer and Limani, Fidan and Usbeck, Ricardo},
  year={2023},
  journal={Gesellschaft f{\"u}r Informatik eV}, 
}

@article{gesese2024nfdi4dso,
  title={NFDI4DSO: Towards a BFO Compliant Ontology for Data Science},
  author={Gesese, Genet Asefa and Waitelonis, J{\"o}rg and Chen, Zongxiong and Schimmler, Sonja and Sack, Harald},
  journal={arXiv preprint arXiv:2408.08698},
  year={2024}
}

@article{codabench,
    title = {Codabench: Flexible, easy-to-use, and reproducible meta-benchmark platform},
    author = {Zhen Xu and Sergio Escalera and Adrien Pavão and Magali Richard and 
                Wei-Wei Tu and Quanming Yao and Huan Zhao and Isabelle Guyon},
    journal = {Patterns},
    volume = {3},
    number = {7},
    pages = {100543},
    year = {2022},
    issn = {2666-3899},
    doi = {https://doi.org/10.1016/j.patter.2022.100543},
}

@inproceedings{ahmad2024forc4cl,
  title={FoRC4CL: a fine-grained field of research classification and annotated dataset of NLP articles},
  author={Ahmad, Raia Abu and Borisova, Ekaterina and Rehm, Georg},
  booktitle={Proceedings of the 2024 Joint International Conference on Computational Linguistics, Language Resources and Evaluation (LREC-COLING 2024)},
  pages={7389--7394},
  year={2024}
}

@inproceedings{abu2024forc,
  title={FoRC@ NSLP2024: Overview and Insights from the Field of Research Classification Shared Task},
  author={Abu Ahmad, Raia and Borisova, Ekaterina and Rehm, Georg},
  booktitle={International Workshop on Natural Scientific Language Processing and Research Knowledge Graphs},
  pages={189--204},
  year={2024},
  organization={Springer}
}

@article{d2025semeval,
  title={SemEval-2025 Task 5: LLMs4Subjects--LLM-based Automated Subject Tagging for a National Technical Library's Open-Access Catalog},
  author={D'Souza, Jennifer and Sadruddin, Sameer and Israel, Holger and Begoin, Mathias and Slawig, Diana},
  journal={arXiv preprint arXiv:2504.07199},
  year={2025}
}

@misc{llms4subjects-dataset,
author = {D'Souza, Jennifer and Sadruddin, Sameer and Israel, Holger and Begoin, Mathias and Slawig, Diana},
doi = {10.5281/zenodo.15185475},
title = {{The SemEval 2025 LLMs4Subjects Shared Task Dataset}},
url = {https://github.com/jd-coderepos/llms4subjects/},
urldate = {2025-07-24},
year = {2024},
}

@InProceedings{homa,
  author    = {Bayrami Asl Tekanlou, Hadi  and  Razmara, Jafar  and  Sanaei, Mahsa  and  Rahgouy, Mostafa  and  Babaei Giglou, Hamed},
  title     = {Homa at SemEval-2025 Task 5: Aligning Librarian Records with OntoAligner for Subject Tagging},
  booktitle      = {Proceedings of the 19th International Workshop on Semantic Evaluation (SemEval-2025)},
  month          = {August},
  year           = {2025},
  address        = {Vienna, Austria},
  publisher      = {Association for Computational Linguistics},
  pages     = {2371--2377},
}

@InProceedings{dnb-ai,
  author    = {Kluge, Lisa  and  Kähler, Maximilian},
  title     = {DNB-AI-Project at SemEval-2025 Task 5: An LLM-Ensemble Approach for Automated Subject Indexing},
  booktitle      = {Proceedings of the 19th International Workshop on Semantic Evaluation (SemEval-2025)},
  month          = {August},
  year           = {2025},
  address        = {Vienna, Austria},
  publisher      = {Association for Computational Linguistics},
  pages     = {1114--1124},
}

@InProceedings{annif,
  author    = {Suominen, Osma  and  Inkinen, Juho  and  Lehtinen, Mona},
  title     = {Annif at SemEval-2025 Task 5: Traditional XMTC augmented by LLMs},
  booktitle      = {Proceedings of the 19th International Workshop on Semantic Evaluation (SemEval-2025)},
  month          = {August},
  year           = {2025},
  address        = {Vienna, Austria},
  publisher      = {Association for Computational Linguistics},
  pages     = {2395--2402},
}

@inproceedings{giglou2024llms4ol,
  title={LLMs4OL 2024 Overview: The 1st Large Language Models for Ontology Learning Challenge},
  author={Giglou, Hamed Babaei and D’Souza, Jennifer and Auer, S{\"o}ren},
  booktitle={Open Conference Proceedings},
  volume={4},
  pages={3--16},
  year={2024}
}

@inproceedings{babaei2023llms4ol,
  title={LLMs4OL: Large language models for ontology learning},
  author={Babaei Giglou, Hamed and D’Souza, Jennifer and Auer, S{\"o}ren},
  booktitle={International Semantic Web Conference},
  pages={408--427},
  year={2023},
  organization={Springer}
}

@article{llms4ol_dataset, 
title={LLMs4OL 2024 Datasets: Toward Ontology Learning with Large Language Models}, volume={4}, DOI={10.52825/ocp.v4i.2480}, journal={Open Conference Proceedings}, author={Babaei Giglou, Hamed and D’Souza, Jennifer and Sadruddin, Sameer and Auer, Sören}, year={2024}, month={Oct.}, pages={17–30} }

@inproceedings{lin-2004-rouge,
    title = "{ROUGE}: A Package for Automatic Evaluation of Summaries",
    author = "Lin, Chin-Yew",
    booktitle = "Text Summarization Branches Out",
    month = jul,
    year = "2004",
    address = "Barcelona, Spain",
    publisher = "Association for Computational Linguistics",
    pages = "74--81"
}

@inproceedings{zhang2020BERTScore,
title={BERTScore: Evaluating Text Generation with BERT},
author={Tianyi Zhang* and Varsha Kishore* and Felix Wu* and Kilian Q. Weinberger and Yoav Artzi},
booktitle={International Conference on Learning Representations},
year={2020},
}

@misc{voorhees2005trec,
  title={TREC: Experiment and evaluation in information retrieval},
  author={Voorhees, EM},
  year={2005},
  publisher={MIT Press}
}

@inproceedings{otto2023gsapnernoveltaskcorpus,
    title = "{GSAP}-{NER}: A Novel Task, Corpus, and Baseline for Scholarly Entity Extraction Focused on Machine Learning Models and Datasets",
    author = {Otto, Wolfgang  and
      Zloch, Matth{\"a}us  and
      Gan, Lu  and
      Karmakar, Saurav  and
      Dietze, Stefan},
    editor = "Bouamor, Houda  and
      Pino, Juan  and
      Bali, Kalika",
    booktitle = "Findings of the Association for Computational Linguistics: EMNLP 2023",
    month = dec,
    year = "2023",
    address = "Singapore",
    publisher = "Association for Computational Linguistics",
    doi = "10.18653/v1/2023.findings-emnlp.548",
    pages = "8166--8176",

}

@article{ragab2023adatime,
  title={Adatime: A benchmarking suite for domain adaptation on time series data},
  author={Ragab, Mohamed and Eldele, Emadeldeen and Tan, Wee Ling and Foo, Chuan-Sheng and Chen, Zhenghua and Wu, Min and Kwoh, Chee-Keong and Li, Xiaoli},
  journal={ACM Transactions on Knowledge Discovery from Data},
  volume={17},
  number={8},
  pages={1--18},
  year={2023},
  publisher={ACM New York, NY}
}

@inproceedings{borisova-scivqa-2025,
  title = "{SciVQA} 2025: Overview of the First Scientific Visual Question Answering Shared Task",
  author = "Borisova, Ekaterina and Rauscher, Nikolas and Rehm, Georg",
  booktitle = "Proceedings of the 5th Workshop on Scholarly  Document Processing (SDP)",
  year = "2025",
  address = "Vienna, Austria",
  comment = "accepted" }

@inproceedings{climatecheck-shared-task,
    title = "The {ClimateCheck} Shared Task: Scientific Fact-Checking of Social Media Claims about Climate Change",
    author = "Abu Ahmad, Raia and Usmanova, Aida and Rehm, Georg",
    booktitle = "Proceedings of the 5th Workshop on Scholarly Document Processing (SDP)",
    year = "2025",
    address = "Vienna, Austria",
}

@inproceedings{somd-2025,
  title = "{SOMD2025}: A Challenging Shared Tasks for Software Related Information Extraction",
  author = "Upadhyaya, Sharmila and Otto, Wolfgang and Krüger, Frank and Dietze, Stefan",
  booktitle = "Proceedings of the 5th Workshop on Scholarly  Document Processing (SDP)",
  year = "2025",
  address = "Vienna, Austria",
  comment = "accepted" }

@inproceedings{francis2025overview,
  author = {Maria Francis and Raia Abu Ahmad and Ekaterina Borisova and
                  Georg Rehm},
  title = {{Overview of the FoRC@NSLP2025 Shared Task: Field of
                  Research Classification for Computational Linguistics and
                  Natural Language Processing Publications}},
  booktitle = {{Joint Proceedings of the ESWC 2025 Workshops and Tutorials
                  co-located with 22nd Extended Semantic Web Conference (ESWC
                  2025)}},
  year = 2025,
  month = 06,
}

@inproceedings{climatecheck-dataset,
    title = "The {ClimateCheck} Dataset: Mapping Social Media Claims About Climate Change to Corresponding Scholarly Articles",
    author = "Abu Ahmad, Raia and Usmanova, Aida and Rehm, Georg",
    booktitle = "Proceedings of the 5th Workshop on Scholarly Document Processing (SDP)",
    year = "2025",
    address = "Vienna, Austria",
    comment = "Accepted for publication" 
}

\end{document}